\documentclass[letterpaper, 10 pt, conference]{ieeeconf}  

\IEEEoverridecommandlockouts                              
 
\usepackage{graphics} 
\usepackage{epsfig} 
\usepackage{amsmath} 
\usepackage{amssymb}
\usepackage{csquotes}
\usepackage{wrapfig}
\usepackage{glossaries}
\usepackage{hyperref}
\usepackage{url}

\newcommand{\rhoexp}{{\rho_{\pi_E}}}
\newcommand{\rhopi}{{\rho_{\pi}}}

\title{\LARGE \bf
Exciting Action: Investigating Efficient Exploration for Learning Musculoskeletal Humanoid Locomotion
}

\author{Henri-Jacques Geiß$^{1, \dagger}$, Firas Al-Hafez$^{1, \ddagger}$, Andre Seyfarth$^{2, \ddagger}$, Jan Peters$^{1, 3, \ddagger}$ and Davide Tateo$^{1, \ddagger}$
\thanks{*This work was supported by the German Science Foundation (DFG) under
grant number SE1042/41-1. Calculations for this research were conducted on the Lichtenberg high-performance computers at TU Darmstadt.}
\thanks{$^{1}$Computer Science Department, Technical University Darmstadt}
\thanks{$^{2}$Human Sciences Department, Technical University Darmstadt}
\thanks{$^{3}$German Research Center for AI (DFKI), Centre for Cognitive Science}
\thanks{$\phantom{^{3}}$Hessian.AI}
\thanks{{$^\dagger$}{\tt\small henri.geiss@uibk.ac.at}}
\thanks{{$^\ddagger$}{\tt\small {name.surname}@tu-darmstadt.de}}%
}

\begin{document}

\newacronym{rl}{RL}{Reinforcement Learning}
\newacronym{irl}{IRL}{Inverse Reinforcement Learning}
\newacronym{il}{IL}{Imitation Learning}
\newacronym{ail}{AIL}{Adversarial Imitation Learning}
\newacronym{le}{LE}{Latent Exploration}
\newacronym{sar}{SAR}{Synergistic Action Representation}

\newacronym{drl}{DRL}{Deep Reinforcement Learning}

\newacronym{avi}{AVI}{Approximate Value-Iteration}
\newacronym{api}{API}{Approximate Policy-Iteration}
\newacronym[plural=MDPs, firstplural=Markov Decision Processes (MDPs)]{mdp}{MDP}{Markov Decision Process}
\newacronym[plural=POMDPs, firstplural=Partially Observable Markov Decision Processes (POMDPs)]{pomdp}{POMDP}{Partially Observable Markov Decision Process}
\newacronym{kl}{KL}{Kullback-Leibler Divergence}
\newacronym{gae}{GAE}{Generalized Advantage Estimation}
\newacronym{trpo}{TRPO}{Trust Region Policy Optimization}
\newacronym{sac}{SAC}{Soft-Actor Critic}
\newacronym{ddpg}{DDPG}{Deep Deterministic Policy Gradient}
\newacronym{td3}{TD3}{Twin Delayed DDPG}
\newacronym{bptt}{BPTT}{Backpropagation Through Time}
\newacronym{s2pg}{S2PG}{Stochastic Stateful Policy Gradient}
\newacronym{pgt}{PGT}{Policy Gradient Theorem}
\newacronym{dpgt}{DPGT}{Deterministic Policy Gradient Theorem}
\newacronym{ppo}{PPO}{Proximal Policy Optimization}
\newacronym{gail}{GAIL}{Generative Adversarial Imitation Learning}
\newacronym{lsiq}{LS-IQ}{Least-Squares Inverse Q-Learning}
\newacronym[plural=LSTMs]{lstm}{LSTM}{Long-Short Term Memory}
\newacronym[plural=GRUs]{gru}{GRU}{Gated Recurrent Units}
\newacronym{rdpg}{RDPG}{Recurrent Deterministic Policy Gradient}
\newacronym{rsvg}{RSVG}{Recurrent Stochastic Value Gradient}
\newacronym{ode}{ODE}{Ordinary Differential Equation}
\newacronym[plural=FFNs, firstplural=Feed-Forward Networks (FFNs)]{ffn}{FFN}{Feed-Forward Network}

\newacronym{rtd3}{RTD3}{Recurrent Twin-Delayed Deep Deterministic Policy Gradient}
\newacronym{rsac}{RSAC}{Recurrent Soft Actor-Critic}

\newacronym[plural=RNNs, firstplural=Recurrent Neural Networks (RNNs)]{rnn}{RNN}{Recurrent Neural Network}

\newacronym[plural=CPGs, firstplural=Central Pattern Generators (CPGs)]{cpg}{CPG}{Central Pattern Generator}
\newacronym{mtus}{MTUs}{Muscle-Tendon Units}
\newacronym{ce}{CE}{Contractile Element}
\newacronym{pe}{PE}{Passive Element}
\glsdisablehyper
\maketitle
\thispagestyle{empty}
\pagestyle{empty}

\begin{abstract}
Learning a locomotion controller for a musculoskeletal system is challenging due to over-actuation and high-dimensional action space. While many reinforcement learning methods attempt to address this issue, they often struggle to learn human-like gaits because of the complexity involved in engineering an effective reward function. In this paper, we demonstrate that adversarial imitation learning can address this issue by analyzing key problems and providing solutions using both current literature and novel techniques. We validate our methodology by learning walking and running gaits on a simulated humanoid model with 16 degrees of freedom and 92 Muscle-Tendon Units, achieving natural-looking gaits with only a few demonstrations. Code is available at \href{https://github.com/henriTUD/musculoco\_learning}{https://github.com/henriTUD/musculoco\_learning}.
\end{abstract}
\IEEEpeerreviewmaketitle

\section{INTRODUCTION}

Locomotion on simulated \emph{musculoskeletal} humanoids requires precise muscle activation patterns. At the same time, it is a vital prerequisite for the design and implementation of gait-assistive devices such as prostheses and exoskeletons. Consequently, the development of biomechanically accurate simulation environments as well as suitable strategies for acquiring control policies for these systems has garnered increasing attention in recent years. Particularly when it comes to applying \gls{rl}, the musculoskeletal domain presents a unique set of challenges.
Firstly, human joints are controlled by multiple antagonistic muscles pulling in different directions, resulting in a high-dimensional action space and significant \emph{over-actuation} in musculoskeletal systems. This not only results in additional functional complexity to learning policies but also severely hampers joint-space exploration, decreasing learning performance.
Moreover, when aiming for distinctively natural and biomechanically accurate locomotion, a well-designed cost function beyond simple sparse rewards becomes crucial. To address this, motion capture joint data have been used to fit reward parameters \cite{schumacher2023natural} or construct trajectory tracking rewards \cite{anand2019deep, peng2017learning}. Many recent studies furthermore directly supply reference joint trajectories to the policies at runtime for increased naturalness \cite{lee2019scalable, park2022generative}. \gls{irl} is a learning paradigm in which a reward function is not handcrafted but learned from expert demonstrations. This approach is particularly well-suited for tasks like walking, where the appropriate reward function is often only understood intuitively \cite{russell1998learning}. Since our goal is to extract a policy using the learned reward function, this learning paradigm is often referred to as apprenticeship learning. In this context, \gls{ail}, where the policy's objective is to generate trajectories indistinguishable from those of the expert, offers a promising framework \cite{orsini2021matters}.
To the best of our knowledge, this principled usage of motion capture data has, to date, not been employed to learn policies for musculoskeletal control of humanoid locomotion with black-box reward function approximation. To establish an initial anchor point for learning such policies with \gls{ail}, we make the following contributions.
(1) We provide the first empirical study on \gls{gail} \cite{ho2016generative} for high-dimensional muscle-actuated humanoid locomotion. 
(2)~Achieving sufficient exploration is a multi-faceted problem, spanning policy, objective function, and the environment. For that reason, we provide an in-depth benchmark over design choices for achieving sufficient exploration and optimal asymptotic performance with \gls{gail}.
In addition to evaluating different policy distributions and exploration objectives, we compare two recent approaches inspired by muscle synergies \cite{chvatal2012voluntary, clark2010merging} in the human nervous system: Latent Exploration \cite{chiappa2024latent} and  \gls{sar}\cite{berg2023sar}. While our study centers on \gls{gail}, its findings are equally applicable to pure \gls{rl} applications.

\begin{figure}
\centering
\includegraphics[scale=0.15]{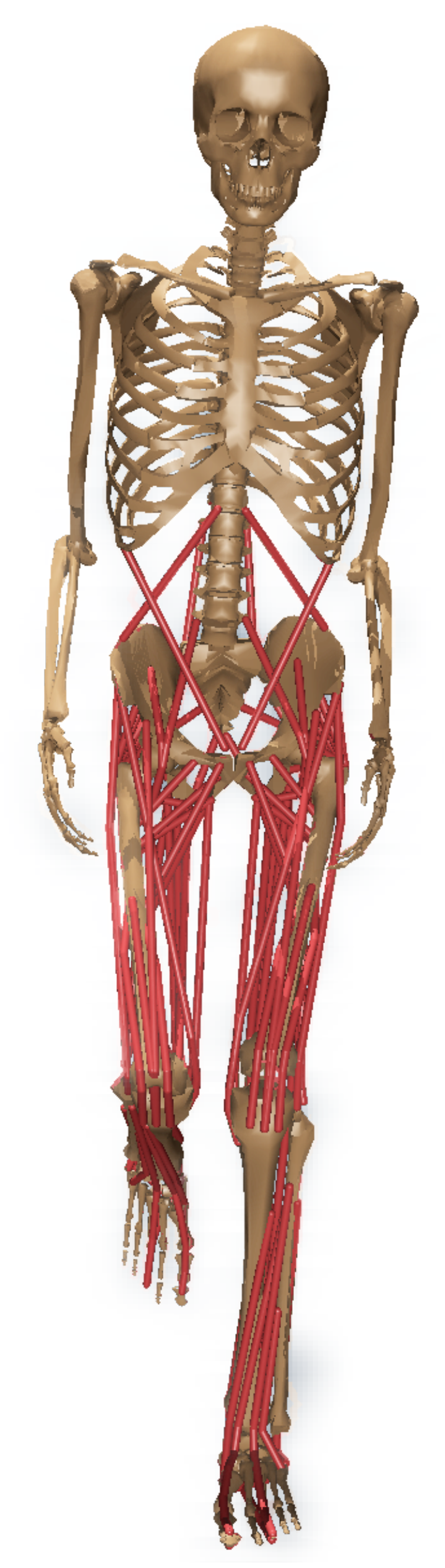}
\includegraphics[scale=0.15]{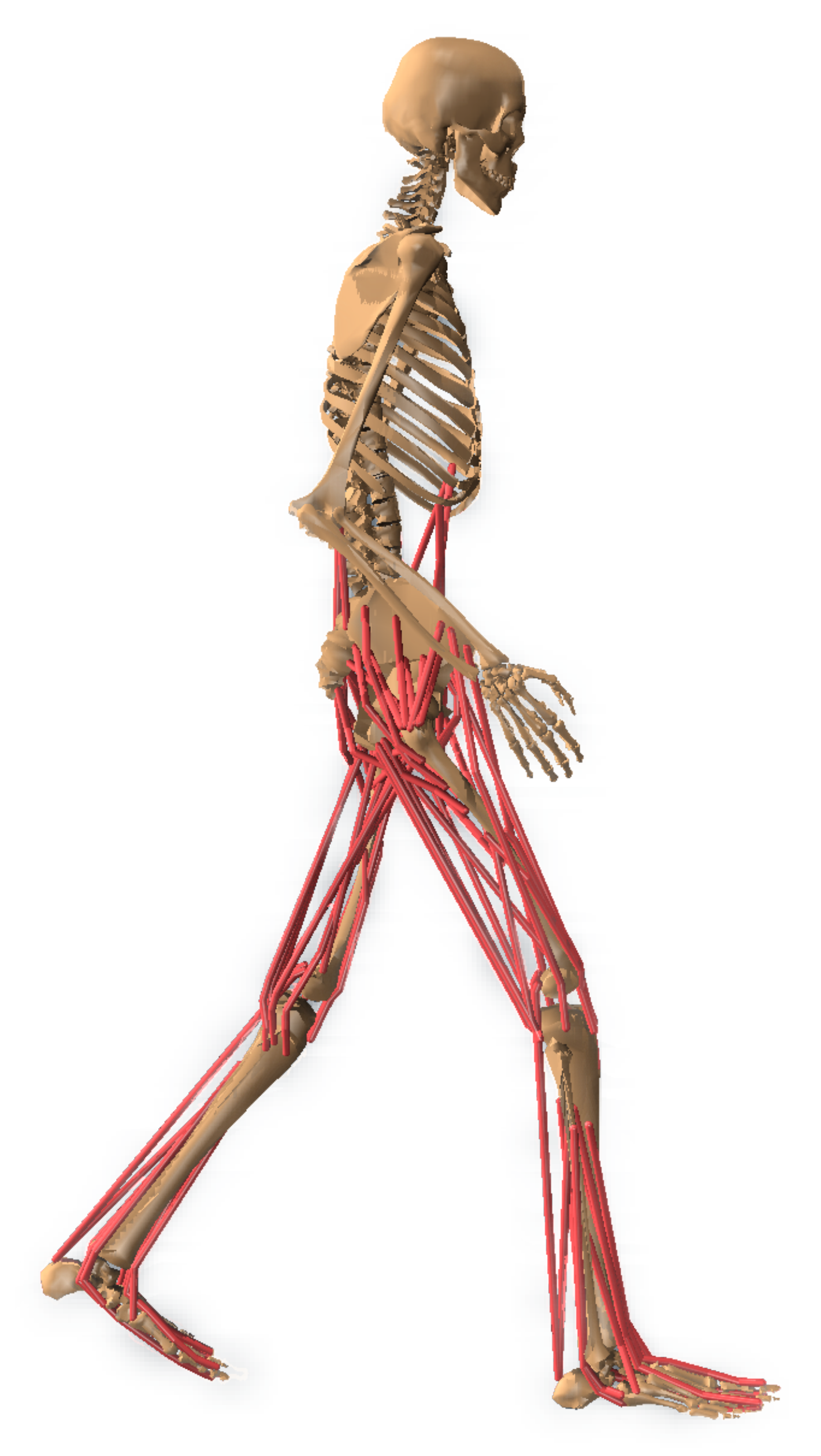}
\caption{\label{figure:thumbnail} Humanoid model with 16 DOFs actuated by 92 Muscle-Tendon Units during running (left) and walking (right).}
\end{figure}


\section{RELATED WORK}
\label{sec:rel_work}
A large amount of research on applying the RL framework for learning muscle-actuated locomotion control was done
as part of challenges at the NeurIPS conferences of the years 2017 - 2019 and 2023 \cite{kidzinski2018learning, kidzinski2020artificial, zhou2019efficient, song2021deep}.
As the humanoid in the first three years of the challenge was actuated by a relatively small amount of muscles, standard end-to-end policies were a feasible option. An Ornstein-Uhlenbeck \cite{uhlenbeck1930theory}  process for generating time-correlated and therefore \emph{colored}~\cite{haunggi1994colored} noise was often employed together with the off-policy DDPG algorithm \cite{kidzinski2018learning, kidzinski2020artificial}.
When employing musculoskeletal models with a higher number of muscles, such approaches lack sufficient exploration capabilties. Consequently, many recent works opt for a two-level network architecture \cite{park2022generative} that decomposes the policy into separate modules.
The first level is responsible for computing desired joint torques \cite{park2022generative, jatesiktat2022muscle} or accelerations \cite{lee2019scalable, qin2022muscle} based on the current state and given reference trajectories,
while the second then predicts the actual muscle excitations from these values.
With such an approach, Lee et al. \cite{lee2019scalable}, for instance, could learn locomotion policies on a humanoid actuated by 346 muscles. In contrast to such specialized approaches, we focus on general exploration schemes, which can be incorporated more easily into any \gls{rl} or \gls{ail} algorithm. 

More related approaches focus on the problem of efficient exploration for RL in the over-actuated regime. One recent advance in that regard, the DEP-RL \cite{schumacher2022dep}, is based on the
notion of \emph{chaining together what changes together}, and hence produces off-policy exploration signals that induce velocity correlated state trajectories as opposed to the stiffening up of traditional over-actuated exploration. While the gait learned with this approach alone pans out rather unnatural and unstable, with additional reward shaping
DEP-RL achieves human-like walking and running behavior on multiple different humanoid simulation environments
\cite{schumacher2023natural}.

Unlike the off-policy DEP-RL approach, recent advancements such as LATTICE \cite{chiappa2024latent} and \gls{sar} \cite{berg2023sar} focus on on-policy synergistic exploration inspired by real muscle synergies. This makes them suitable for \gls{ail}, which is why both methods will undergo empirical evaluation in this study and will be discussed in detail later on.
Regarding the usage of motion capture data for learning policies that are especially biomechanically accurate, such datasets have been used for
the above mentioned reference trajectories that are given as input to the policy at runtime \cite{lee2019scalable, qin2022muscle, jatesiktat2022muscle}, for naive trajectory
mimicking reward functions \cite{peng2017learning, lee2019scalable, de2021deep, anand2019deep} as well as fitting of joint range reward coefficients \cite{schumacher2023natural}.
That being said, principled algorithms of \gls{ail} have not been used for this task in prior work, a research gap we aim to close with this work.

While many different instances of \gls{ail} exist \cite{ho2016generative, peng2018variational, garg2021, al2022ls}, we observed that \gls{gail} performs most reliably when trained on expert data derived from noisy motion capture data without direct access to actions.

\section{BACKGROUND}
\label{sec:background}

\subsection{Muscle Modelling}
In close to all cases in the literature, muscle fibres and the respective tendons they are attached to are computationally modeled together as \enquote{massless linear actuators with active and passive properties} \cite{rajagopal2016full} considered as \gls{mtus}. 
\begin{wrapfigure}{r}{0.18\textwidth}
\centering
\includegraphics[scale=0.55]{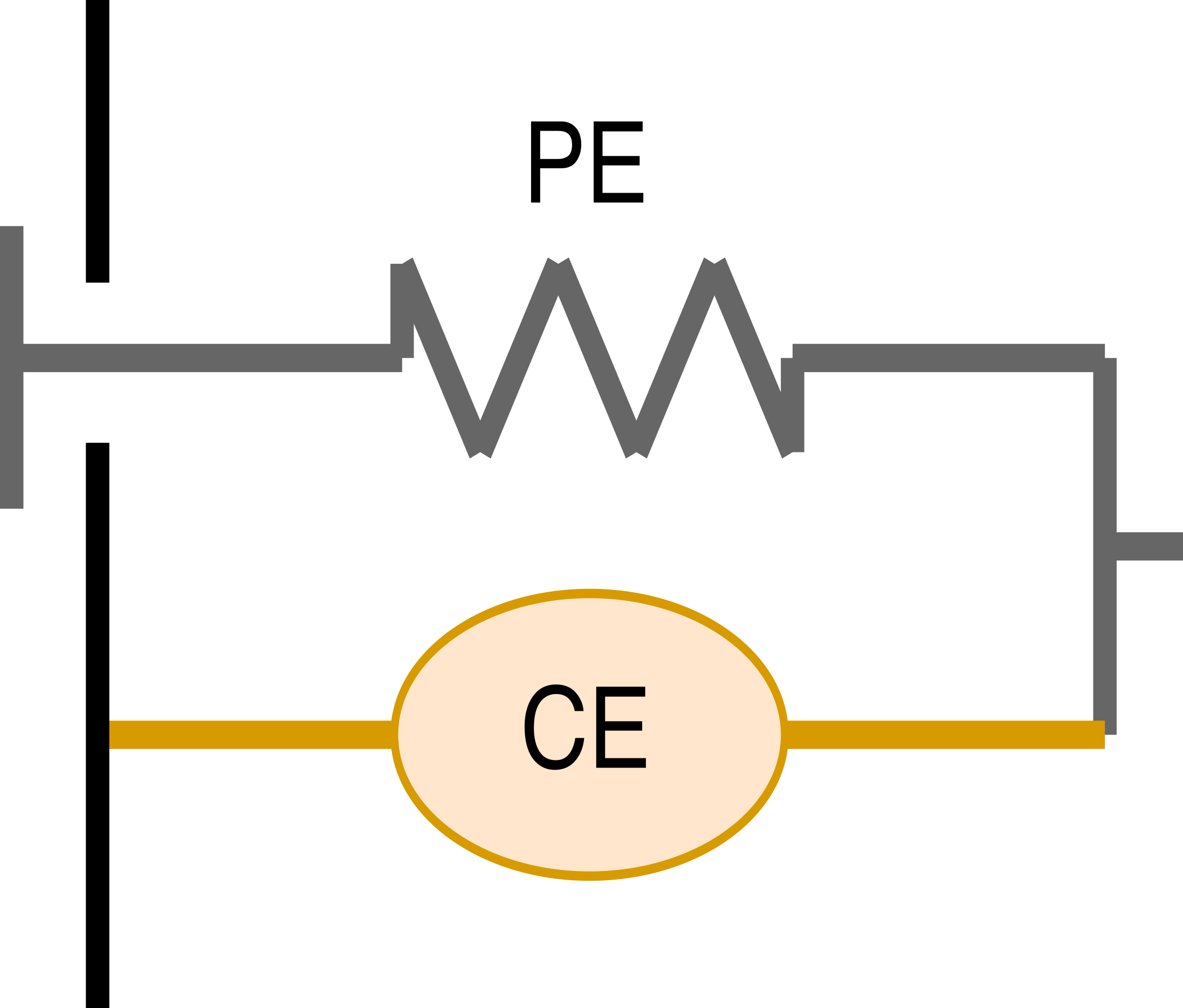}
\caption{\label{fig:mtu} Basic composition of a Muscle-Tendon Unit}
\end{wrapfigure}
While the precise setup of the modeled force applicants, springs and dampers varies between studies, at their core, all variants contain a \gls{ce} and a \gls{pe} as displayed in Fig. \ref{fig:mtu} \cite{geyer2010muscle, rajagopal2016full}. Moreover, it is assumed that the force generated by the \gls{mtus} is only dependent on the activation, length, and velocity of the \gls{ce} resulting in the FLV function. In MuJoCo, the simulation framework of choice in this work, the FLV function is defined as the following \cite{todorov2012mujoco}: 
\begin{equation*}
\textrm{FLV(L,V, z)} = F_{L}(L) \cdot F_{V}(V) \cdot \textrm{z} + F_P(L).
\end{equation*}
It combines the active force generated by the \gls{ce} given its current length, velocity, and activation with the passive force generated by the \gls{pe} at lengths larger than the optimal resting length $L_0$. 
The action emitted by the \gls{rl} policy does not directly control the muscle activation $\textrm{z}$ in the FLV function. Instead, $\textrm{z}$ is derived through a first-order nonlinear filter, with Mujoco-internal parameters $\tau_{\textrm{act}}$ and $\tau_{\textrm{deact}}$,  where the control signal \emph{a} is clamped into $[0, 1]$ \cite{todorov2012mujoco}
\begin{equation*}
\label{eq:filter_dynamics}
\frac{\partial}{\partial t} \textrm{z} = \frac{\textrm{a} - \textrm{z}}{\tau(\textrm{a}, \textrm{z})}
\end{equation*}
\begin{equation*}
\tau(\textrm{a}, \textrm{z}) = \begin{cases}
			\tau_{\textrm{act}}(0.5+1.5 \cdot \textrm{z}), & \textrm{a} - \textrm{z} > 0\\
            \frac{\tau_{\textrm{deact}}}{0.5+1.5\cdot\textrm{z}}, & \textrm{a} - \textrm{z} \leq 0 
		 \end{cases} \, . 
\end{equation*}
It is crucial to understand that additional smoothing in an over-actuated action space adds complexity to exploration while clamping actions within an environment can also impede learning, as discussed later on.

\subsection{Musculoskeletal Humanoid}
With a computational muscle model at hand, the next step involves integrating it with a rigid body humanoid skeleton model using OpenSim \cite{delp2007opensim}, a major simulation framework widely used today. Models in OpenSim are defined using XML files that specify the kinematic structure of the humanoid and the parameters and attachment points of the \gls{mtus} that actuate it. Recently, to better integrate with existing \gls{rl} libraries, simulation environments have converted prominent OpenSim models into XML format for the Mujoco physics simulator \cite{al2023locomujoco, caggiano2022myosuite}, albeit with some sacrifice in biomechanical accuracy. We here will apply the Mujoco-converted version of a humanoid initially compiled by Hamner et al. \cite{hamner2010muscle} as part of the LocoMujoco \cite{al2023locomujoco} library. Displayed in Fig. \ref{figure:thumbnail}, the environment's 16 degrees of freedom are actuated by 92 MTUs. Table \ref{tab:lower_limb_movements} contains a subset of the system's muscles and the joint movement that they induce when excited. The large degree of over-actuation of the human lower limb apparatus becomes apparent, especially when considering that some of the muscles are controlled even more fine-grained in our model. For instance, the Gluteus Maximus is separated into three distinct MTUs in the action space.
Naturally, this suggests a high degree of correlation of muscle activations over the gait cycle. And in fact, prior studies based on Electromyography (EMG) measurements of muscle activity during gait, and human motor control in general, suggest that the human nervous system exhibits neural pathways for \enquote{modular control of muscles in groups} \cite{chvatal2012voluntary} by which it achieves muscle synergies~\cite{clark2010merging}. 
As an example, the authors in \cite{clark2010merging} could explain at least 85\% of variance in muscle activity with four modules, for instance, combining Soleus and Gastrocnemius or Tibialis Anterior and Rectus Femoris.



\begin{table}[t!]
    \vspace{5pt}
    \centering
       \caption{\label{tab:lower_limb_movements} Lower Limb Joint Actuation}
   \begin{tabular}{| c | l |}
   \hline
        \textbf{Joint Movement} & \textbf{Muscles Involved} \\
        \hline
        \hline
        Plantar Flexion & Gastrocnemius, Soleus \\
        \hline
        Dorsiflexion & Tibialis Anterior, Extensor Hallucis Longus,\\
       & Extensor Digitorum Longus\\
        \hline
        \hline
        Knee Extension & Quadriceps Femoris, Rectus Femoris,\\ 
        & Vastus Muscles \\
        \hline
        Knee Flexion & Biceps Femoris, Semitendinosus, \\ 
        & Semimembranosus, Gastrocnemius \\
        \hline
        \hline
        Hip Flexion & Iliopsoas, Rectus Femoris, Sartorius \\
        \hline
        Hip Extension & Gluteus Maximus, Biceps Femoris, \\
        & Semitendinosus, Semimembranosus \\
        \hline
        \hline
        Hip Abduction & Gluteus Medius, Gluteus Minimus\\
        \hline
        Hip Adduction & Adductor Longus, Adductor Brevis, \\ 
       & Adductor Magnus, Gracilis \\
        \hline
        \hline
        Hip Internal Rotation & Gluteus Medius, Gluteus Minimus\\
        \hline
        Hip External Rotation & Piriformis, Gemelli \\
        \hline
    \end{tabular}
    \vspace{-2em}
\end{table}
\subsection{Problem Formulation}
We model the environment as a \gls{mdp}. \gls{mdp}s are described by a tuple $(\mathcal{S}, \mathcal{A}, P, r, \gamma, \mu_0 )$,
with state space $\mathcal{S}$ and action space $\mathcal{A}$.
The stochastic dynamics of the environment $P: \mathcal{S} \times \mathcal{A} \times \mathcal{S} \rightarrow \mathbb{R}^+$ encodes how the environment state changes when a certain action is applied. To encode a certain task that the agent should learn, we define the reward function $r: \mathcal{S} \times \mathcal{A} \rightarrow \mathbb{R}$, with a discount factor $\gamma$. One episode then consists of a certain amount of iterations in which
the agent observes a state $s \in \mathcal{S}$ of the environment, predicts an action $a \in \mathcal{A}$ using the policy $\pi:\mathcal{S} \times \mathcal{A} \rightarrow \mathbb{R}^+$, that induces a transition with probability $P(s'|s, a)$ into the next state $s' \in \mathcal{S}$, where it receives the reward $r(s, a)$. We define an occupancy measure $\rhopi (s, a) = \pi(a|s) \sum_{t=0}^\infty \gamma^t \mu^\pi_t(s)$, where $\mu_{t}^\pi(s')=\int_{s,a}\mu^\pi_{t}(s)\pi(a|s)P(s'|s,a)da\,ds$ is the state distribution for $t>0$, with the initial state distribution $\mu^\pi_0(s)=\mu_0(s)$. The occupancy measure allows us to denote the expected reward under policy $\pi$ as $\mathbb{E}_{\rhopi}[ r(s,a)] \triangleq \mathbb{E}[\sum_{t=0}^\infty \gamma^t r(s_t,a_t)]$, where $s_0 \sim \mu_0$, $a_t \sim \pi(.|s_t)$ and $s_{t+1} \sim P(.|s_t, a_t)$ for $t>0$.

\subsection{Adversarial Imitation Learning} 
As mentioned before, we use the \gls{gail} algorithm. The latter optimizes the following saddle point problem
\begin{equation}
      \inf_{D} \sup_{\pi} -\mathbb{E}_{\rhoexp}[\log (D(s,a))] - \mathbb{E}_{\rhopi}[\log(1- D(s,a))]\,,\label{ref:log_reg}
\end{equation}
where $D \in (0,1)^{S \times A}$ is the discriminator trained to output values close to $1$ under the occupancy of the expert $\rhoexp$ and values close to $0$ under the occupancy of the policy $\rhopi$. Note that the inner optimization over policies constitutes a standard \gls{rl} problem, in which the policy optimizes the reward $-\log(1- D(s,a))$. In practice, parameterized models for the discriminator and the policy are used, and the inner and outer loop in Eq. \eqref{ref:log_reg} are not fully solved at each iteration, but only one gradient step is taken alternately. We use the \gls{trpo} algorithm \cite{schulman2015trust} to optimize the inner \gls{rl} loop. Typically, to ensure that the policy sufficiently explores in the environment, an entropy bonus $\mathcal{H}(\pi)$ is used together with the TRPO surrogate loss, given an advantage estimator $A(s, a)$ and transitions sampled from a policy $q$
\begin{equation}
 \mathcal{L}_{\textrm{TRPO}} = \mathbb{E}_{s,a \sim q, \mathcal{P}} \left[\frac{\pi(a|s)}{q(a|s)}{A}(s, a)\right] + \lambda\mathcal{H}(\pi).  \label{eq:ent_bonus}
\end{equation}
Because expert actions are unavailable, we modify the discriminator to use states alone, omitting actions as demonstrated in \cite{torabi2019_GaifO}.

\section{METHODS}
\label{sec:method}
Typical \gls{rl} methods often overlook the unique characteristics of muscle-actuated systems. In this section, we identify two key issues with standard \gls{rl} approaches that lead to performance problems in muscle-actuated systems.

\textit{1.) Impact of Clamped Actions on Policy Gradients}: We hypothesize that actions clamped by the environment not only eliminate policy gradients for out-of-bounds actions but pose a particular risk in musculoskeletal models, especially when paired with common exploration objectives like the one in Eq. \ref{eq:ent_bonus}. The absence of policy gradients from the environment leaves the entropy bonus as the sole objective for out-of-bounds actions. This can result in high entropy throughout training, preventing convergence to a near-deterministic policy. We propose to tackle these issues from multiple angles. In Section \ref{sec:pi_dist}, we introduce policy distributions with bounded support, which inherently prevent sampling out-of-bounds actions. Alternatively, in Section \ref{sec:expl_obj}, we propose incorporating additional objectives directly into the policy objective to generate gradients for out-of-bounds actions that counterbalance entropy maximization. 

\textit{2.) Neglecting Muscle Synergies:}
As noted earlier, the \gls{mtus} in our model, much like real muscles, can only exert torque in one direction, are often inactive for significant portions of the gait cycle, and exhibit highly correlated action trajectories due to muscle synergies. These characteristics are typically overlooked in traditional \gls{rl}, leading us to hypothesize that the absence of synergies is a key factor contributing to poor performance. In section, \ref{sec:musc_syn}, we present specialized methods for learning synergies.
\subsection{Policy Distribution} \label{sec:pi_dist}
\paragraph{Unbounded Spherical Gaussian} The standard policy distribution for most on-policy \gls{rl} algorithms is an unbounded Gaussian distribution. Here, the mean $\mu_{\theta}(s)$ is predicted by the policy network, while the standard deviation is parameterized state-independent with a single scalar parameter $\sigma_{\theta}$ for each action dimension, resulting in a spherical multivariate normal distribution where each dimension's density is
\begin{equation*}
\label{eq:uni_gaussian}
\pi(a|s) = \frac{1}{\sqrt{2\pi}\sigma_{\theta}}\exp\left(-\frac{(a - \mu_{\theta}(s))^2}{2\sigma_{\theta}^2}\right)\,. 
\end{equation*}
Since this distribution is unbounded, it suffers from the aforementioned issues of clamped actions spaces with exploration objectives. This distribution requires additional objectives as discussed later in Section \ref{sec:expl_obj}.
\paragraph{Squashed Gaussian} A common method to enforce a probability density with bounded support is \emph{tanh-squashing}~\cite{haarnoja2018soft}. Here, an action $u$ is sampled from an unbounded density $\mu(u|s)$ and then forced into the action range by applying the element-wise hyperbolic tangent $a = \tanh(u)$. By virtue of the change of variables formula, a closed-form solution for the log-likelihood of the sampled action can be derived \cite{haarnoja2018soft}
\begin{equation*}
\log \pi(a|s) = \log \mu(u|s) - \sum_{i=1}^{D}\log(1 - \tanh^2(u_i))\,.
\end{equation*}
With large enough initial standard deviations, the squashed Gaussian additionally provides a bi-modal distribution for exploration, that might be better suited for the properties of muscle activation signals laid out above. Since this distribution is bounded, action clamping has no effect.

\paragraph{Beta Distribution} The other type of distribution with bounded support that we will probe for exploration in this work is the Beta distribution. It is defined on the interval $[0, 1]$
by two positive parameters $\alpha$ and $\beta$, yielding the following density function
\begin{equation*}
f(x; \alpha, \beta) = \frac{\Gamma(\alpha - \beta)}{\Gamma(\alpha)\Gamma(\beta)}x^{\alpha-1}(1-x)^{\beta-1}
\end{equation*}
\begin{equation*}
\Gamma(n) = (n-1)!\, . 
\end{equation*}
Previous uses of the Beta distribution in RL have enforced unimodality of the distribution by adding 1 to the strictly positive policy network outputs $\alpha_{\theta}(s)$ and $\beta_{\theta}(s)$ \cite{chou2017improving}, which we will also probe in this study. 
Moreover, the Beta distribution can also be parameterized from its mean, that is predicted by the policy network and standard deviation. The values for $\alpha$ and $\beta$ can then be derived as follows
\begin{equation*}
\alpha = \left( \frac{1 - \mu}{\sigma^2} - \frac{1}{\mu}\right)\mu^2, \quad \beta = \alpha\left(\frac{1}{\mu}-1\right)\,.
\end{equation*}
For the Beta distribution in general it has to hold that $\sigma^2 < \mu(1-\mu)$. Moreover, a tighter bound on $\sigma$ can be derived that on top ensures unimodality of the resulting distribution
\begin{equation*}
\label{eq:max_std_beta}
 \sigma^2 < \mu \min\left(\frac{\mu(1-\mu)}{1+\mu}, \frac{(1 - \mu)^2}{2-\mu}  \right)\,.
\end{equation*}
\subsection{Exploration Objective}
\label{sec:expl_obj}
As mentioned before, unbounded policy distributions require additional objectives to address the problem of clamped actions with entropy maximization. More precisely, based on the observation that \gls{mtus} are fully \enquote{relaxed}, i.e. have zero activation, during joint movements in the opposite direction of their attachment \cite{schumacher2023natural}, the entropy maximization objective can be expected to actually cause decreased coverage of the effective action space. Concretely, in preliminary test runs with the default entropy objective of GAIL with an unbounded spherical Gaussian distribution we observe that the policy resorts to almost bang-bang control of the \gls{mtus}. Hereby, it can move the predicted mean arbitrarily far out-of-bounds of the effective action space of the environment, while not reducing or even increasing its standard deviation and therefore its entropy. Consequently, for tasks that are solvable with bang-bang-style action signals, the entropy objective leads to effectively \emph{decreased} exploration.

\begin{wrapfigure}{l}{0.25\textwidth}
\includegraphics[scale=0.31]{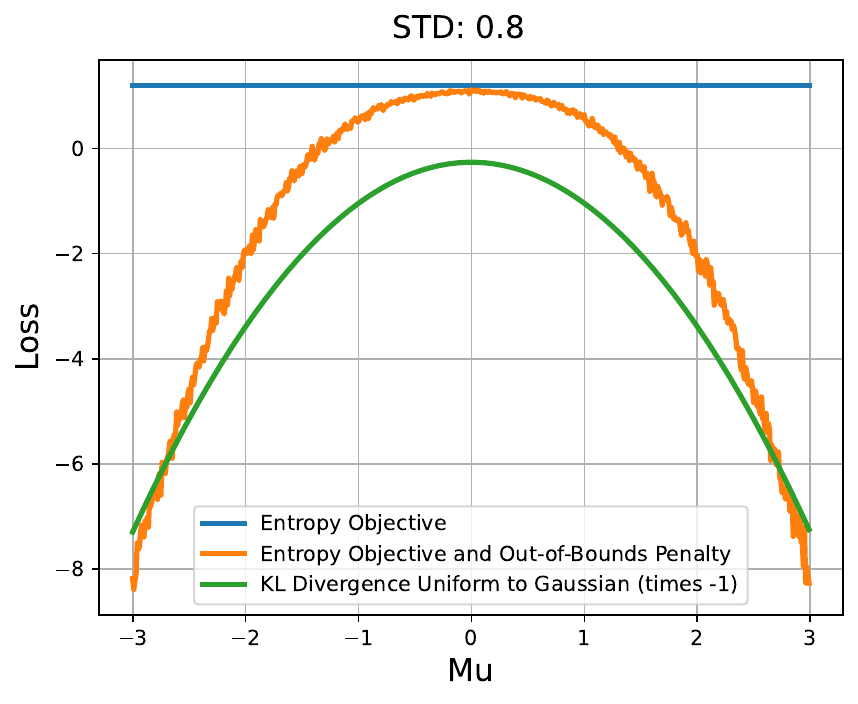}
\caption{\label{fig:ent_conf} Comparison of exploration objectives depending on the mean of the gaussian.}
\end{wrapfigure}

\paragraph{Target Entropy} We probe replacing the entropy objective with a target entropy as a naive solution for this problem. With a target entropy value $h_{\textrm{target}}$ the overall objective changes to
\begin{equation*}
\mathcal{L}_{\textrm{TE}} = \mathbb{E}_{s,a \sim q, \mathcal{P}} \left[\frac{\pi(a|s)}{q(a|s)}A(s, a)\right] -\lambda_{\textrm{TE}}(\mathcal{H}(\pi)-h_{\textrm{target}})^2.
\end{equation*}
\paragraph{Flipped Uniform KL Objective} While a target entropy in the case of a Gaussian distribution only influences the standard deviation parameter, we also present an exploration objective that affects the mean of the distribution. As established in prior works, for bounded distributions, maximizing entropy is the exact dual to maximizing the \gls{kl} to a uniform distribution.
Since this operation is not defined for an unbounded distribution support, we propose the \emph{Flipped Uniform KL Objective} for musculoskeletal exploration
\begin{equation*}
\mathcal{L}_{\textrm{FKL}} = \mathcal{L}_{\textrm{TRPO}} - \lambda_{\textrm{FKL}} \mathbb{E}_{s \sim q, \mathcal{P}} \left[D_{KL}(\mathcal{U}||\pi(s))\right].
\end{equation*}
We derive KL divergence of a uniform distribution on the action space to a multivariate gaussian to be the following
with density of the uniform distribution $u$ and the bounds $a_i$ and $b_i$ of the action range, shifted by the mean of the gaussian distribution for the respective dimension
\begin{equation*}
\begin{aligned}
D_{KL}&(\mathcal{U}||\mathcal{N}) = \log(u) + \frac{1}{2}\log|\Sigma| + \frac{|\mathcal{A}|}{2}\log(2\pi) \\ & + \frac{1}{2}u\left[\sum_p(\prod_{i \neq p} (b_i - a_i)) \Sigma^{-1}_{pp} \frac{(b_p^3 - a_p^3)}{3}\right] \\ & +  \frac{1}{2}u \left[ \sum_{p \neq q}(\prod_{i \neq p, q} (b_i - a_i)) \Sigma^{-1}_{pq} \frac{(b_p^2 - a_p^2)(b_q^2 - a_q^2)}{4}\right] \,.
\end{aligned}
\end{equation*}
This penalty pushes the mean of the unbounded Gaussian distribution into the bounds by introducing a policy gradient for out-of-bounds actions. 
\paragraph{Out-of-Bounds Penalty}
Similarly, to incentivize policies to place significant amounts of their probability mass within the bounds of the action space for arbitrary unbounded distributions, we chose to combine the \gls{gail} discriminator reward with an environment reward. Specifically, as demonstrated in prior works \cite{furuta2021co}, penalizing high absolute scalar action values forces the policy to balance imitation performance with maintaining low mean and standard deviation.
After preliminary testing of different options for the design of this reward function, we opt for a \emph{squared out-of-bounds penalty}. It is calculated like the following for action dimension $i$, a lower action bound $b$, an upper bound $c$ and a scaling coefficient $\beta$
\begin{align*}
 r_t = \beta\sum_{i=0} ^{|\mathcal{A}|}r_{t,i},
& &
 r_{t,i} = 
\begin{cases}
-1(a_{t,i} - b)^2, & \text{if} \quad a_{t,i} < b\\
-1(a_{t,i} - c)^2, & \text{if} \quad a_{t,i} > c\\
0,  & \text{otherwise} 
\end{cases}\,.
\end{align*}
Fig. \ref{fig:ent_conf} compares the loss functions of the standard entropy objective, with the ones resulting from the flipped uniform KL objective and the out-of-bounds penalty together with the entropy objective. 
Both of the proposed methods clearly put a direct gradient towards the center of the action space on the predicted mean. The combination out-of-bounds penalty and entropy objective has the additional property, that the standard deviation decreases significantly as the mean moves further towards the bounds of the action space.
\subsection{Muscle Synergies}\label{sec:musc_syn}
As stated in section \ref{sec:rel_work}, several methodologies have been presented in prior work to make use of the muscle synergies for efficient exploration, two of which we will also test for \gls{ail} in this study.

\paragraph{Synergistic Action Representation}
Based on the \emph{correlation} of muscle activation signals that is predicted by muscle synergies, \gls{sar}, as recently presented Berg et al. \cite{berg2023sar}, directly employs classic methods for dimensionality reduction to represent the action space on a lower-dimensional and synergistic manifold. In the case of acquiring this representation on the same task as the policy learning is supposed to take place, they split training into two phases~\cite{berg2023sar}. The \emph{Play Phase} lasts for a pre-determined amount of environment steps, in which an arbitrary RL policy is trained on the task with the original over-actuated actions space $\mathcal{A}$. 
Now, after collecting actions $M_{act} \in \mathbb{R}^{|\mathcal{A}|\times t}$ of this policy for $t$ simulation steps, a lower dimensional representation of the action space for this task can be acquired with traditional means of dimensionality reduction. For that, Berg et al. \cite{berg2023sar} make use of ICA, PCA and a normalization: 
\begin{equation*}
 SAR := |T_{ICA}(T_{PCA}(M_{act}))|.
\end{equation*}
Given this representation, in the subsequent \emph{Training Phase}, a completely new policy is learned, that now acts in this $N_{syn}$-dimensional synergistic space. Similar to what can be said about the Spinal Cord during real human locomotion, movement is now controlled and \emph{explored} on the level of muscle groups instead of distinct \gls{mtus}. In its full extent, SAR constructs an action space of size $|\mathcal{A}| + N_{syn}$ that offers both, the original and the synergistic pathways. That being said, Berg et al. \cite{berg2023sar} found that the synergistic manifold alone suffices for control of musculoskeletal locomotion, hence we will also rely on it here.

\paragraph{Latent Exploration}
The other approach for synergistic exploration of musculoskeletal actions that we will probe in this study is Latent Exploration, which was established as part of the LATTICE framework by Chiappa et al.~\cite{chiappa2024latent}. 
While their full methodology encapsulates latent, state-dependent and time-correlated exploration, we will focus on Latent Exploration in this work.
The \enquote{latent state} here refers to the last hidden layer state $x$ of the policy network based on which the mean of the action is calculate linearly before adding spherical gaussian noise with a state-independent parameter $\Sigma_a$
\begin{equation*}
   a = W_xx + \epsilon \quad \epsilon \sim \mathcal{N}(0, \Sigma_a).
\end{equation*}
For Latent Exploration, we add $N_x$ (the size of the latent state) parameters to our policy, which represent the
variance of each dimension of the latent space and therefore now sample with a second uncorrelated Gaussian distribution
$\hat{x} \sim \mathcal{N}(x, \Sigma_x)$
\begin{equation*}
   a = W_x\hat{x} + \epsilon, \quad \hat{x} \sim \mathcal{N}(x, \Sigma_x), \quad \epsilon \sim \mathcal{N}(0, \Sigma_a).
\end{equation*}
With simple transformations, we can therefore express $a$ as being sampled
from a single distribution in pure Latent Exploration
\begin{equation*}
    \label{eq:purely_latent}
   a \sim \mathcal{N}(W_x x, \Sigma_a + W_x\Sigma_xW_x^T).
\end{equation*}
With this construction of the covariance matrix, the more similar the linear weights in $W_x$  for two dimensions of the action space are, the higher their covariance in the multivariate Gaussian will be. As muscles that pull in the same direction on the same joint can be expected to have similar weights in $W_x$, this leads to an increasingly synergistic sampling of agonist muscles with ongoing updates to $W_x$. For all experiments in this work, weights $W_x$ were detached from the gradient graph when computing the covariance matrix, which was beneficial in preliminary testing.

\section{EXPERIMENTS}
\begin{figure*}[ht!]
\vspace{5pt}
\centering
\includegraphics[scale=0.85]{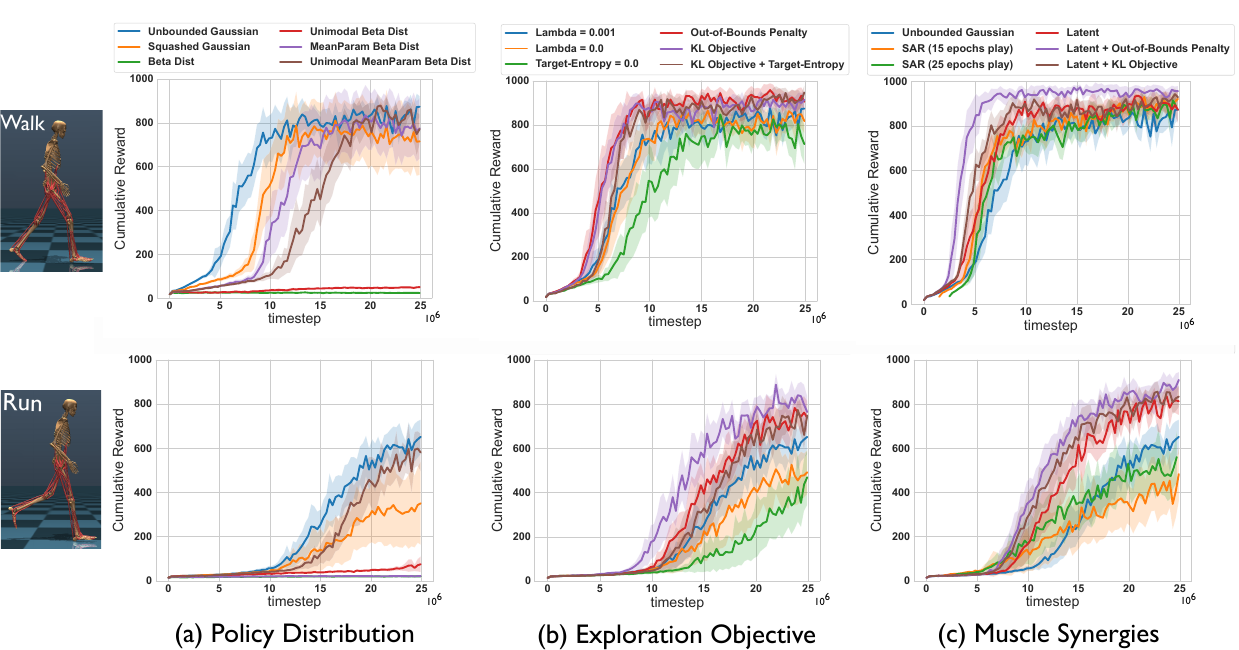}
\caption{\label{fig:exp_res} Target velocity performance curves for imitation learning of \emph{walking} and \emph{running} with the different (a) policy distributions (b) exploration objectives and (c) methods for synergistic exploration introduced in section \ref{sec:method}. All experiments were started with 15 seeds.}
\vspace{-1em}
\end{figure*}
We conduct a thorough comparison of the presented methods for learning musculoskeletal locomotion control in two tasks, \emph{walking} at 1.25 $\frac{m}{s}$ and \emph{running} at 2.5 $\frac{m}{s}$. For both of these, we use the motion capture datasets provided by LocoMujoco \cite{al2023locomujoco}, which contain recorded trajectories of one average-height healthy male subject, performing locomotion at constant speed on a treadmill, as the expert dataset \cite{al2023locomujoco}. 
For the humanoid, we chose to set the \enquote{use\_box\_feet} option and exclude the torque controlled arms from the action space. The state space, consisting of angles and angular velocities for all DOFs, as well as the height and translational velocities of the center of mass, amounts to a dimensionality of 36. All of the three function approximators - policy, critic and discriminator - were constituted by MLPs with two hidden layers of size 512 and 256 and \emph{identity} last layer activation.

We define a reward function for the environment that is a suitable scalar-valued compression of the performance of the agent in the task which it is learning by imitation. Here, we decide on “target velocity” as a suitable measure. For each policy control step it returns the exponential negative square of the difference of the velocity of the center of mass of the humanoid into the x-direction with the target velocity. Importantly, “target velocity” is never included into the actual loss function, instead it indicates how similar the agent is performing compared to the imitation dataset and how many timesteps the average episode lasts before reaching an absorbing state. We set a maximum episode step amount of 1000, inducing the same value as the maximally achievable target velocity reward.
Fig. \ref{fig:exp_res} displays the results of our study, separated by whether policies were trained for walking or running and which category was ablated. 
\subsection{Policy Distribution}
Fig. \ref{fig:exp_res}.a compares three different choices for the underlying type of probability distribution of the policy. For the Beta distribution we additionally distinct between enforcing unimodality and whether the policy network emits alpha and beta values or the mean of the distribution.

\textbf{Results:} We find that using a bounded probability distribution for the policy does not induce performance improvements, as in both tasks the unbounded diagonal Gaussian distribution yields the best performance. 
Generally, we find that \emph{running} is a significantly harder task to learn than walking. While target-velocity rewards of around 800 can be reached in around 20 million environment steps with the unbounded and squashed Gaussian as well as the mean parameterized Beta distribution, for running only an asymptotic reward of 600 can be observed at maximum. Even with further hyper-parameter search, the traditional parameterization of the Beta distribution did not show any adequate performance.
\subsection{Exploration Objective}
To validate our hypothesis that the standard maximum entropy objective is inadequate in the muscle-actuated domain, we compare it to the \emph{flipped uniform KL objective} and \emph{out-of-bounds penalty} established above. Additionally, we probe a setup without any additional loss, a target entropy of zero and the KL objective combined with target entropy.

\begin{figure}
\vspace{5pt}
\centering
\includegraphics[scale=0.5]{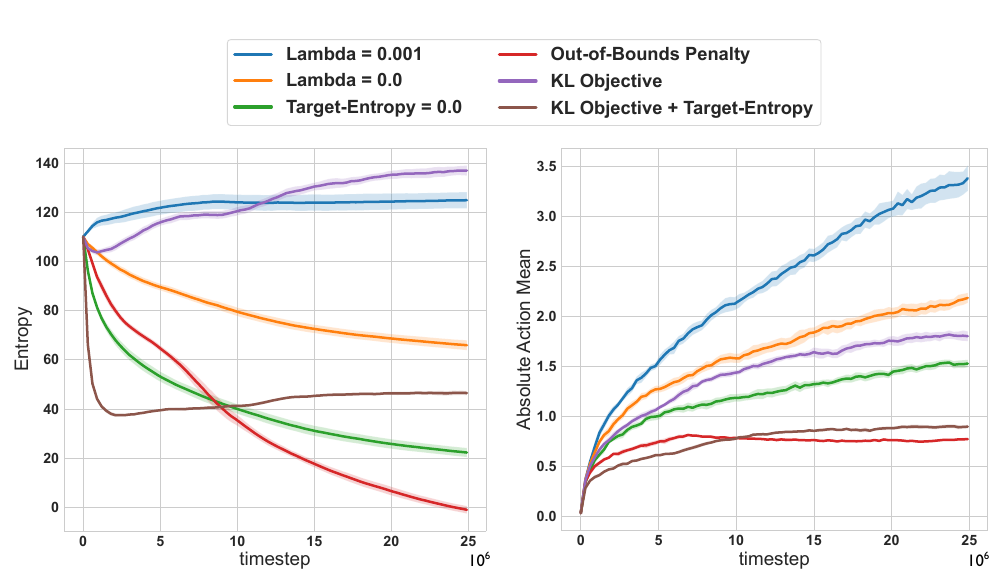}
\caption{\label{fig:ent+abs} Development of the policy distribution's entropy and absolute action mean under the different exploration objective setups.}
\vspace{-1em}
\end{figure}

\textbf{Results:}
Additionally to performance curves in Fig. \ref{fig:exp_res}.b, here we also capture the respective entropy curves as well as how the magnitude of the action mean evolves (see Fig. \ref{fig:ent+abs}). With them we can experimentally validate what was theorized in section \ref{sec:expl_obj}. Since a \emph{bang-bang} style action signal on the bounds of the action range does lead to stable locomotion with reasonable imitation of the expert dataset, setting a entropy objective coefficient of 0.001 does lead to a continually increasing action mean and a stable entropy. Both, setting said coefficient to zero and replacing it with a target entropy, do lead to a decrease in entropy over the course of the experiment and consequently also lower absolute action means. Yet, a performance increase can not be observed for either setup. In contrast, the other three objectives studied in this experiment - the out-of-bounds penalty, the KL objective and the KL objective combined with a target entropy - all directly entail a regularization of the predicted mean of the emitted action and all lead to significant improvements in asymptotic performance. In the case of \emph{walking}, the out-of-bounds penalty and the pure flipped uniform KL objective outperform the baseline setup equally, while for \emph{running} the latter yields the steepest incline and best final performance, improving over the standard objective by 200 cumulative reward. Considering that the entropy trajectory of the \enquote{target entropy} setup develops very similar to the one of the out-of-bounds penalty, while its absolute action mean is below the one of the KL objective, we can conclude from this experiment that a direct regularization of the predicted mean significantly improves performance for an unbounded gaussian policy distribution.
\subsection{Synergistic Exploration}
Lastly, we probe the two recently introduced approaches Latent Exploration and SAR for synergistic on-policy exploration of muscle clusters. As Latent Exploration yields an unbounded multivariate Gaussian, we additionally probe it in combination with the out-of-bounds penalty and the flipped uniform KL objective. For SAR we also use the out-of-bounds penalty in the \emph{play phase} of the experiments. Studying whether SAR can be advantageous in terms of speed of learning or asymptotic performance, we fit ICAPCA after 1.5 and 2.5 million environment steps respectively.

\begin{figure}
\vspace{5pt}
\centering
\includegraphics[scale=0.55]{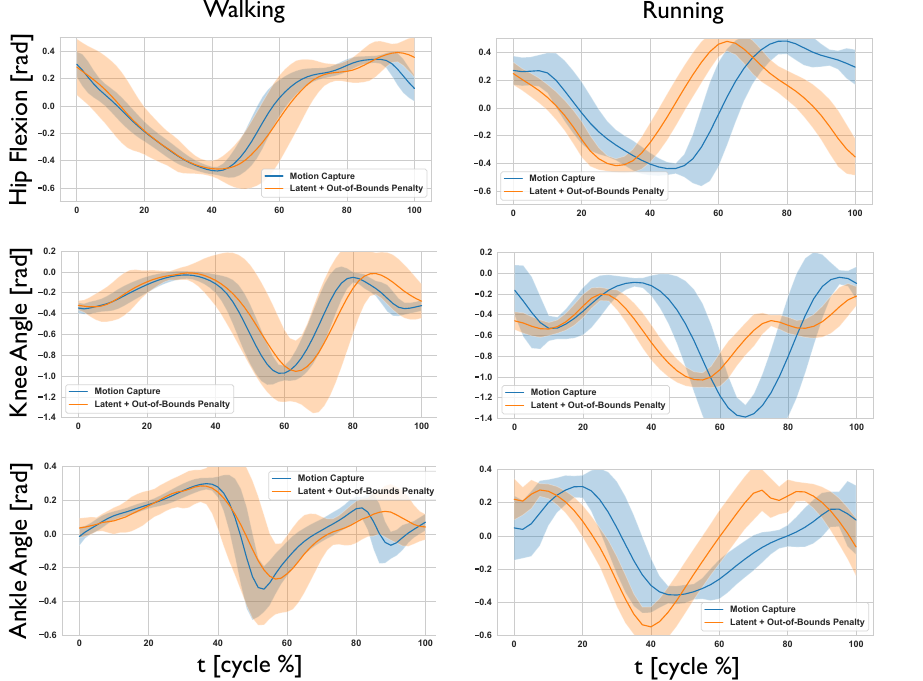}
\caption{\label{fig:mocap} Comparison of the trajectories of the motion capture dataset with the policy learned with latent exploration and an out-of-bounds penalty. For \emph{walking} this setup achieves near-optimal imitation of the expert.}
\vspace{-1em}
\end{figure}

\textbf{Results:}
For \emph{walking}, both, SAR and Latent Exploration, outperform the unbounded Gaussian asymptotically. A very significant improvement over this baseline can be observed for the combination of Latent Exploration with the out-of-bounds penalty. With this setup 800 target velocity reward is reached in 5 million timesteps, with a final performance of 970 that is close to the maximal possible value. Regarding \emph{running}, only the three variants of Latent Exploration that were probed had a positive effect on the performance, with the out-of-bounds penalty exploration objective again yielding the best result, making it the overall best-performing setup in our study. To have a white view of the learned gaits, we compare their joint trajectories to the motion capture dataset, shown in Fig. \ref{fig:mocap}. For \emph{walking}, as suggested by the near-optimal reward, the policy learns to imitate the expert dataset very closely. In contrast for \emph{running}, while the general patterns align, the policy learns a faster gait in comparison to the expert and correspondingly completes more than one gait cycle for each cycle of the motion capture data.
\section{CONCLUSION}
Using the GAIL algorithm and a 16 DOF humanoid with 92 MTUs we performed an empirical investigation on three key design choices to achieve sufficient exploration in the muscle- and therefore over-actuated domain.
Regarding the policy distribution, applying the bounded squashed Gaussian or Beta distribution did not pay dividends, yet for the latter, we find that a parameterization through the mean and enforced unimodality can have a positive effect on learning performance. Furthermore, when using an unbounded Gaussian, we find that a direct regularization of the mean through the introduced flipped uniform KL objective or the out-of-bounds penalty is crucial for preventing the policy from moving the mean out of bounds of the action range and causing clear performance gains. The best-performing setup, achieving near-optimal performance and alignment with the motion capture data, uses Latent Exploration for synergistic exploration of agonistic muscles together with said out-of-bounds penalty. This work is an initial anchor point for progress on achieving maximal biomechanical accuracy in musculoskeletal control through the usage of IL and the incorporation of natural principles like \emph{muscle synergies} in the learning framework, which in turn could result in new insights on human locomotion and improvements to the design of gait assistive devices.

\clearpage
\bibliographystyle{plain}
\bibliography{lit}

\end{document}